\documentclass[conference]{IEEEtran}
\IEEEoverridecommandlockouts
\usepackage{cite}
\usepackage{amsmath,amssymb,amsfonts}
\usepackage{algorithmic}
\usepackage{graphicx}
\usepackage{textcomp}
\usepackage{xcolor}

\usepackage{support-caption}
\usepackage[font=small,labelfont=bf]{caption}
\usepackage{mathtools}
\usepackage{booktabs}

\DeclareMathOperator*{\argmax}{argmax}

\usepackage{hyperref}
\hypersetup{
    colorlinks,
    linkcolor={red!50!black},
    citecolor={blue!50!black},
    urlcolor={blue!50!black}
}

\def\BibTeX{{\rm B\kern-.05em{\sc i\kern-.025em b}\kern-.08em
    T\kern-.1667em\lower.7ex\hbox{E}\kern-.125emX}}
\begin{document}

\title{Learning Efficient Unsupervised Satellite Image-based Building Damage Detection}

\author{\IEEEauthorblockN{Yiyun Zhang, Zijian Wang, Yadan Luo, Xin Yu, Zi Huang}
\IEEEauthorblockA{
% \textit{Data Science Discipline, School of Electrical Engineering and Computer Science} \\
\textit{The University of Queensland} \\
Brisbane, Australia \\
\{yiyun.zhang, zijian.wang, y.luo, xin.yu\}@uq.edu.au, huang@itee.uq.edu.au
}}

% \author{\IEEEauthorblockN{Yiyun Zhang, Zijian Wang, Yadan Luo, Xin Yu, Zi Huang}
% \IEEEauthorblockA{The University of Queensland, Brisbane, Australia\\
% \{yiyun.zhang, zijian.wang, y.luo, xin.yu\}@uq.edu.au, huang@itee.uq.edu.au
% }}

\maketitle

\newcommand{\ie}{\textit{i.e.}}
\newcommand{\eg}{\textit{e.g.}}
\newcommand{\etal}{\textit{et al.}}
\newcommand{\etc}{\textit{etc}}

\begin{abstract}
Existing Building Damage Detection (BDD) methods always require labour-intensive pixel-level annotations of buildings and their conditions, hence largely limiting their applications. In this paper, we investigate a challenging yet practical scenario of BDD, Unsupervised Building Damage Detection (U-BDD), where only unlabelled pre- and post-disaster satellite image pairs are provided. As a pilot study, we have first proposed an advanced U-BDD baseline that leverages pre-trained vision-language foundation models (\ie, Grounding DINO, SAM and CLIP) to address the U-BDD task. However, the apparent domain gap between satellite and generic images causes low confidence in the foundation models used to identify buildings and their damages. In response, we further present a novel self-supervised framework, U-BDD++, which improves upon the U-BDD baseline by addressing domain-specific issues associated with satellite imagery. Furthermore, the new Building Proposal Generation (BPG) module and the CLIP-enabled noisy Building Proposal Selection (CLIP-BPS) module in U-BDD++ ensure high-quality self-training. Extensive experiments on the widely used building damage assessment benchmark demonstrate the effectiveness of the proposed method for unsupervised building damage detection. The presented annotation-free and foundation model-based paradigm ensures an efficient learning phase. This study opens a new direction for real-world BDD and sets a strong baseline for future research. The code is available at \href{https://github.com/fzmi/ubdd}{https://github.com/fzmi/ubdd}.
\end{abstract}

% Fig 1: Cover Image
\begin{figure}[!t]
    \centering
    \includegraphics[width=0.485\textwidth]{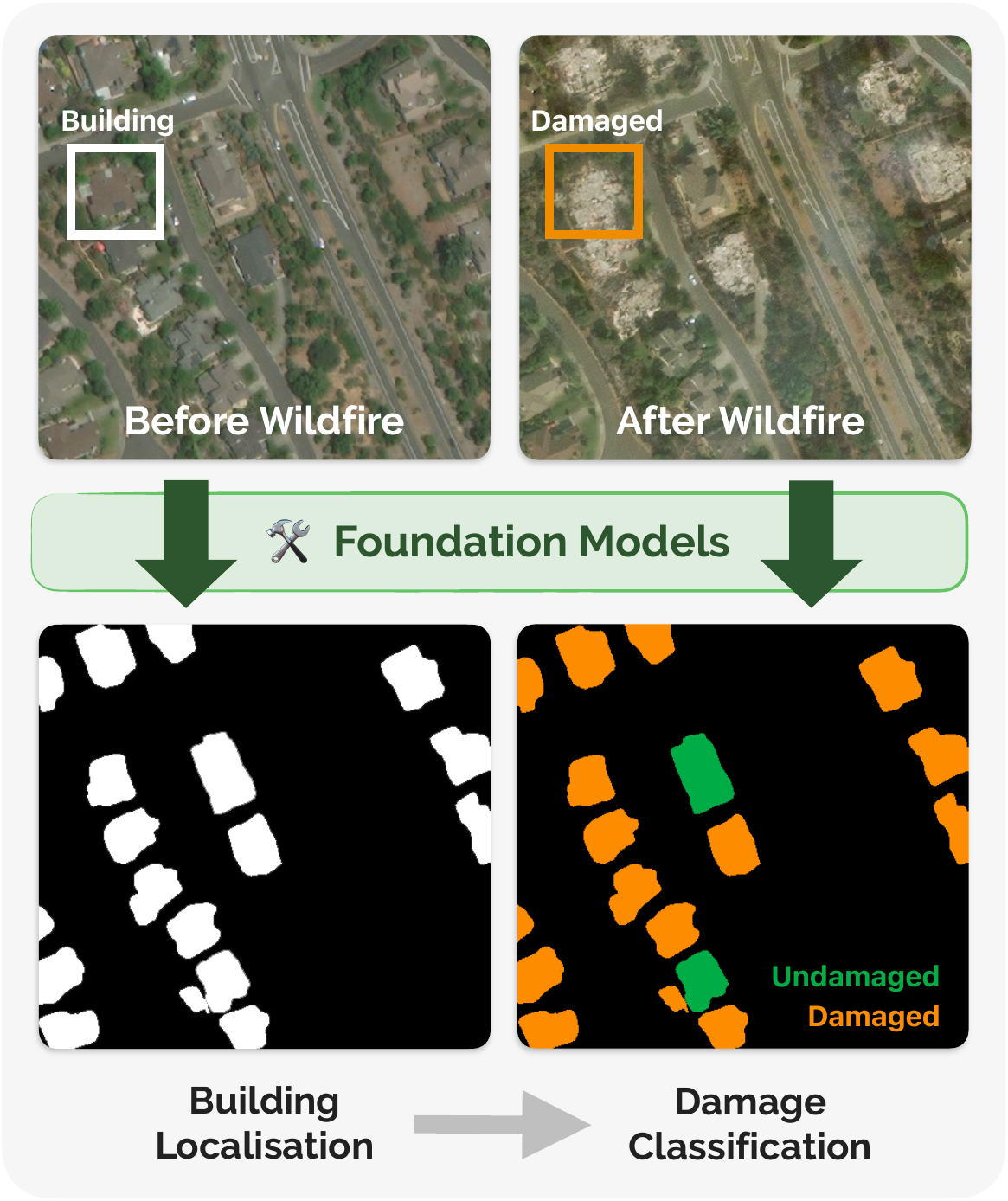}
    \caption{\textbf{High-level illustration of a U-BDD approach.} Provided with a pre-disaster satellite image (top-left) and the corresponding post-disaster image (top-right) in an area of interest, pre-trained foundation models can be applied to perform building localisation (bottom-left) and damage classification (bottom-right) masks.}
    \label{fig:Cover}
\end{figure}


\section{Introduction}
\label{sec:Introduction}

Natural disasters pose significant threats to individuals and communities worldwide, necessitating swift and accurate post-disaster evaluation. One of the essential procedures for post-disaster relief organisations is to identify damaged buildings in the affected areas, which is crucial for coordinating the optimal response and disaster recovery. 
Although traditional on-site surveys are accurate, they can be hazardous for the ground assessment personnel and time-consuming. Fortunately, the recent advancements in remote sensing and deep learning have paved the way for more efficient assessment of building damage, eliminating the need for physical inspection.

Existing Building Damage Detection (BDD) algorithms~\cite{bddmtf, bdd-unet, bdanet, ppm-ssnet} focus on determining if buildings have been damaged from bi-temporal satellite image pairs (\ie, pre-disaster and post-disaster satellite images of the same geolocation). Previous methods treat BDD as a segmentation task, in which a classification task is performed to assess damages (\eg, damaged or undamaged). However, these methods heavily rely on comprehensive pixel-level annotations of building masks, which are time-consuming and labour-intensive. It would be preferable to detect damages with minimal human labelling efforts from the readily available bi-temporal satellite images.

\begin{figure*}[ht]
    \centering
    \includegraphics[width=1.0\textwidth]{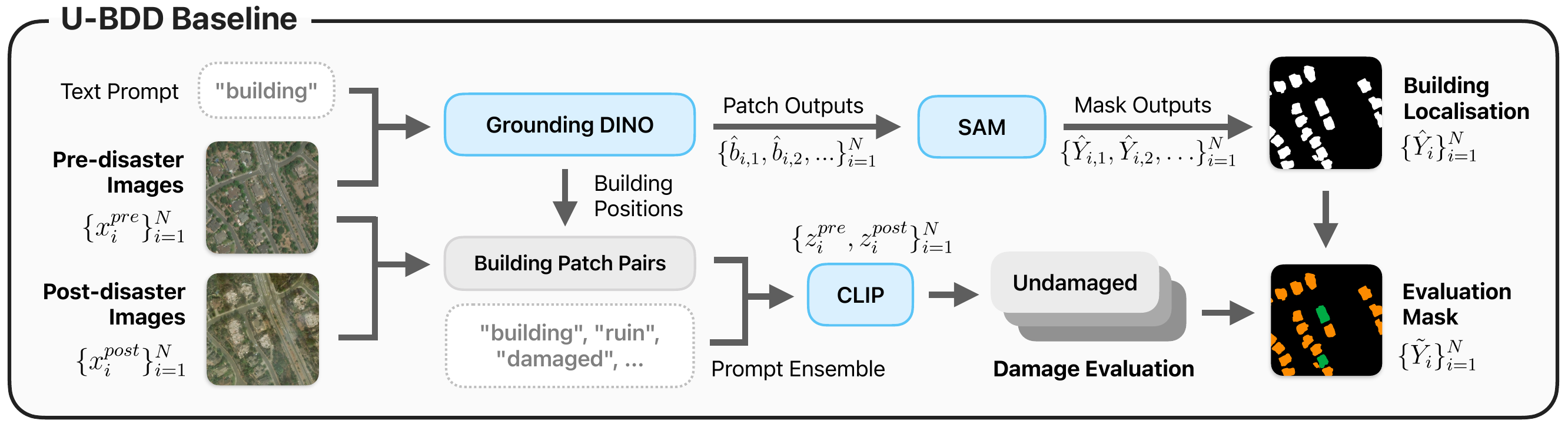}
    \captionsetup{width=0.96\linewidth}
    \caption{\textbf{The schema of the U-BDD baseline}. The pre-disaster images are processed through Grounding DINO and SAM, yielding a building segmentation mask. Concurrently, pre- and post-disaster image pairs, along with the bounding boxes predicted from Grounding DINO, are passed to the CLIP model for damage assessment per building. The final evaluation mask is obtained by integrating the building segmentation mask with the damage predictions.}
    \label{fig:Baseline}
\end{figure*}

In this paper, we are motivated to develop an innovative yet practical building damage detection method, named \textbf{Unsupervised Building Damage Detection} (U-BDD). Under this setting, only unlabelled pre-disaster and post-disaster satellite image pairs are provided. The absence of manual annotations imposes considerable difficulties in extracting building and damage features from the image pairs, leading to a unique challenge to BDD. To the best of our knowledge, none of the existing BDD methods could possibly work in this scenario. Therefore, we design an advanced baseline method by leveraging the strong zero-shot inference capacity of vision-language foundation models. The high-level idea is visualised in Fig.~\ref{fig:Cover}. The proposed advanced U-BDD baseline employs three foundation models, including Grounding DINO~\cite{gdino}, CLIP~\cite{clip} and SAM~\cite{sam}, in a two-stage process to locate potential buildings and identify their respective conditions, as depicted in Fig.~\ref{fig:Baseline}.

Based on our empirical study, we observe that the U-BDD baseline is constrained by considerable domain gaps between the pre-trained data of the foundation models and satellite imagery. Direct application of a pre-trained foundation model without fine-tuning can lead to suboptimal predictions, particularly in building localisation, with the probability of misidentifying unrelated objects as buildings. Such existence of the domain gap is mainly because: 1) The resolution of satellite images is limited, which can cause buildings to appear as small as just a few pixels. This makes accurate detection a challenging task. 2) The top-down perspective inherent in satellite imagery can result in objects with similar semantic characteristics, such as parking lots and swimming pools. These are man-made structures, but not strictly buildings. As a consequence, the models exhibit reduced discriminative confidence, leading to potential misclassification of objects and less reliable results.

In light of these challenges, we propose a robust framework, \textbf{U-BDD++}, which is designed to ameliorate domain-specific issues associated with satellite imagery. U-BDD++ builds upon the two-stage structure of the U-BDD baseline and utilises fine-tuned models for both building localisation and damage classification. The comprehensive framework is illustrated in Fig.~\ref{fig:U-BDD++}. The fine-tuned models are trained in a self-supervised manner, guided by pseudo-labels generated by foundation models. In the building localisation stage, we employ fine-tuned pre-trained detection networks, such as DINO~\cite{dino} and other DETRs~\cite{li2022dn,liu2022dabdetr}, to obtain more accurate bounding boxes of buildings. The preliminary pseudo-labels for the model supervision are produced by our two novel building proposal processing modules, namely the \textbf{Building Proposal Generation (BPG)} module and the \textbf{CLIP-enabled noisy Building Proposal Selection (CLIP-BPS)} module. The BPG module addresses the resolution challenge by incorporating multiscale image transformations and resolution enhancement techniques. This approach allows Grounding DINO~\cite{gdino} or other pre-trained foundation models to have adequate bounding box proposals in various sizes. Meanwhile, the CLIP-BPS module tackles the issue of semantically similar objects by using the discriminative capabilities of the CLIP~\cite{clip} model. This method effectively minimises false positives in building detection, thus improving the quality of the pseudo-labels and contributing significantly to the enhancement of damage classification. In the damage classification stage, we adopt a self-supervised training strategy that involves CLIP in the generation of pseudo-labels for different types of damage, facilitating self-fine-tuning. The fine-tuned CLIP model used in this stage is specifically designed for classification, thereby promoting enhanced training. This sophisticated framework addresses the challenges of the domain gap between satellite imagery and pre-trained models, paving the way for a more efficient and effective approach to the U-BDD task. In summary, the key contributions of our work are as follows:

\begin{itemize}
    \item We define and introduce an innovative yet practical Unsupervised Building Damage Detection (U-BDD) task, which does not rely on costly and laborious pixel-wise annotations of buildings. We design a foundation model-empowered baseline that aims to address this challenging setting by exploiting the zero-shot inference capability.

    \item We further present U-BDD++, a novel self-supervised framework that improves upon the U-BDD baseline by addressing domain-specific issues associated with satellite imagery. The proposed Building Proposal Generation (BPG) module and the CLIP-enabled noisy Building Proposal Selection (CLIP-BPS) module ensure high-quality pseudo-label generations.
    
    \item The proposed U-BDD++ framework has proven to dramatically enhance the performance of our baseline. Specifically, U-BDD++ surpasses the U-BDD baseline by a large margin of 18.3\% and 8.1\% on the F1 metrics in building localisation and damage classification respectively, which verify the effectiveness of our proposed framework modules.
\end{itemize}

% Fig 2: U-BDD Baseline

\section{Related Work}
\label{sec:RelatedWork}

\textbf{Disaster Assessments.} In the field of disaster assessment, the analysis of building damage is critical, encompassing detection, segmentation, and damage assessment. \cite{bddcnn} employs change detection features produced by a convolutional neural network to analyse the building damage. \cite{bdanet} divides the task into two sub-tasks, first localising buildings and then assessing their damage degrees.

Although prior works~\cite{tilon2020post,ji2018identifying} use post-disaster images alone, it can be challenging to obtain precise boundaries or contours of complete building objects. Therefore, using a pair of satellite images for building damage assessment is more reliable, as shown in~\cite{bdanet}. Another stream of works simplifies the damage assessment task as binary classification, while others differentiate building damage levels, such as minor or significant damages.

Researchers often use a Siamese semantic segmentation network for building localisation and damage assessment, such as U-Net~\cite{unet}, along with feature fusion techniques, including feature concatenation~\cite{bddmtf}, difference calculation~\cite{rescuenet}, or attention mechanisms~\cite{bdanet}. The segmentation networks are trained on bi-temporal satellite images with pixel-level labels, either by multitasking strategy~\cite{bddmtf} or by dividing the problem into two stages: building localisation and damage classification~\cite{bdanet}.

Furthermore, Shen \etal~\cite{bdanet-2} introduced a two-stage convolutional neural network, in which they use UNet to extract the localisation of buildings in the first stage and classify damages with a two-branch multiscale classification network in the second stage. Furthermore, \cite{bdanet-2} also designs a bilateral attention mechanism to exchange information between pre- and post-disaster images. Zheng \etal~\cite{zheng2021building} output accurate building locations via a localisation network instead of a segmentation network. Wu \etal~\cite{bdd-unet} adopted attention mechanisms into U-Net, which are implemented with different backbones to extract pre- and post-disaster image features for damage detection. Bai \etal~\cite{ppm-ssnet} employed residual blocks with the dilated convolutional operation to enlarge receptive fields and channel attention (\ie, squeeze-and-excitation~\cite{se}) to enhance feature representations. Thus, their method achieves better building segmentation performance, facilitating damage assessment. However, all methods require a large amount of manual annotation of the data, making them not practical in real situations due to timely responses and limited resources.

\medskip
\textbf{Pre-trained Foundation Models.}
In recent years, large-scale text pre-training on transformer-based models has become increasingly popular for improving visual understanding. This has led to the development of vision-language models (VLM) and has shown impressive performance in various visual understanding tasks. Radford \etal~\cite{clip} proposed Contrastive Language-Image Pre-Training (CLIP) to model the relationship between language and vision. Then, CLIP can use language prompts for zero-shot prediction tasks.

Unlike traditional computer vision models that rely solely on visual features, CLIP combines visual and textual information to make predictions about the content of images. CLIP is trained on a vast corpus of images and text data, enabling it to learn complex visual and linguistic concepts. The model is pre-trained on a diverse set of tasks, including image classification, captioning, and question-answering, without requiring any task-specific training data. This allows CLIP to perform well on a wide range of downstream tasks, including object detection, image retrieval, and visual question answering, among others. Li \etal~\cite{glip} solved the detection limitation by proposing bounding boxes for object candidates and performing cross-attention. Kirillov \etal proposed the Segment Anything Model (SAM)~\cite{sam} can also transfer zero-shot to new image distributions and tasks. Liu \etal~\cite{gdino} developed an open-set object detector, dubbed Grounding DINO, by combining Transformer-based detector DINO~\cite{dino} with grounded pre-training. It can detect arbitrary objects with user-specified text.

\section{U-BDD: A Baseline Approach}
\label{sec:Baseline}

\subsection{Overview}
\label{subsec:U-BDDDefinition}

\textbf{U-BDD Definition.} Consider $X=\{x^{pre}_i, x^{post}_i\}_{i=1}^N$ as a set of $N$ bi-temporal pairs of satellite images (\ie, pre- and post-disaster images). Each image pair $\{x^{pre}_i, x^{post}_i\} \in \mathbb{R}^{C \times H \times W}$ has $C$ channels, and spatial dimensions of height $H$ and width $W$. It is assumed that the geolocations of the pre- and post-disaster image pair are aligned. Let $Y=\{y_i\}_{i=1}^{N}$ denote the corresponding pixel-wise ground-truth labels for damage evaluation. Commonly, each segmentation mask $y_i \in \{0, 1, 2\}^{1\times H \times W}$, where $0, 1, 2$ represent ``background'', ``undamaged building'', and ``damaged building'', respectively. The purpose of U-BDD is to derive a model $M$ through learning from $X$ exclusively, such that $M$ provides a mapping from the image domain $X$ to the output label domain $Y$.

Given the absence of existing solutions apt for the U-BDD task, our first initiative was the creation of a novel, two-stage unsupervised building damage detection approach, denoted as \textbf{U-BDD baseline}. The overall pipeline of the U-BDD baseline is illustrated in Fig.~\ref{fig:Baseline}. The workflow consists of building localisation (explained in Section \ref{subsec:BuildingLocalisation}) and damage classification (discussed in Section \ref{subsec:DamageClassification}). Both of these stages exploit the zero-shot inference capabilities of pre-trained foundation models, including Grounding DINO~\cite{gdino}, CLIP~\cite{clip} and SAM~\cite{sam}. The final evaluation masks $Y$ are generated by combining the building outputs from the localisation stage with the damage assessment outputs derived from the classification stage.

\subsection{Building Localisation}
\label{subsec:BuildingLocalisation}

At the building localisation stage, the U-BDD baseline exploits pre-trained Grounding DINO~\cite{gdino} and SAM~\cite{sam} models to predict a segmentation mask $\hat{Y}_i \in \{0, 1\}^{1\times H \times W}$, where $0$ denotes ``background'', and $1$ denotes ``building''. Notably, at this stage, no damage evaluation is conducted. Since disasters can result in buildings being completely destroyed or buried under lava or floodwater, only pre-disaster images are used to identify the original locations of the buildings.

Grounding DINO~\cite{gdino} processes each pre-disaster image $x_{i}^{pre}$, and uses ``building'' as the text prompt. It initially generates a set of predicted building bounding boxes, each associated with a confidence score. The model then applies a box threshold $\sigma_{G}$ to remove bounding boxes with confidences below $\sigma_{G}$, outputting a final set of $J_i$ bounding boxes $\hat{B}_i = \{\hat{b}_{i,j}\}_{j=1}^{J_i}$. The building bounding boxes for each image $\hat{B}_i$ are used as prompts for SAM~\cite{sam}. After calculating the image embedding for each pre-disaster image, SAM~\cite{sam} performs box prediction for each bounding box $\hat{b}_{i, j}$, and returns a binary segmentation mask
$\hat{Y}_{i, j}\in\{0, 1\}^{H\times W}$,
% H x W is correct here
indicating a more precise outline of the building. To get a final building localisation mask $\hat{Y}_{i}$ for the whole image $x^{pre}_i$, we merge all building segmentation masks of the image $\{\hat{Y}_{i, j}\}^{J_i}_{j = 1}$ pixel-wise with the largest value.

\subsection{Damage Classification}
\label{subsec:DamageClassification}

At the damage classification stage, the U-BDD baseline utilises CLIP~\cite{clip} for zero-shot inference to classify if buildings predicted by Grounding DINO are damaged. Since relying solely on post-disaster images for classification may overlook the crucial changes between pre-disaster and post-disaster scenes, we use bi-temporal inputs for CLIP to conduct damage classification. Specifically, for each bounding box generated by Grounding DINO in the previous stage $\hat{b}_{i, j}$, we obtain the corresponding patch pair $\{x^{pre}_{i, j}, x^{post}_{i, j}\}$ from both pre- and post-disaster images. These patch pairs then serve as image inputs for CLIP. For the text prompts, we adopt a prompt ensemble strategy for CLIP to generate probabilities for both a set of ``positive'' prompts (denoted as $Pt^{+}$) and a set of ``negative'' prompts (denoted as $Pt^{-}$). A ``positive'' prompt is normally a phrase that conveys the building is not damaged, like ``a satellite photo of a building'', and ``normal building''. Similarly, a ``negative'' prompt is a phrase that implies the building is damaged, such as ``damaged building'', or simply ``ruin''. The purpose of prompt ensemble is to make the prediction more robust by allowing the CLIP model to compare not just the positive- and negative-prompt probabilities of the post-disaster image, but also the change in positive-prompt confidence between pre- and post-disaster images.

Let $E_I$ and $E_T$ denote the image encoder and text encoder of CLIP, respectively. Using these encoders, we compute the logits for the image pair $\{x^{pre}_{i, j}, x^{post}_{i, j}\}$ and the prompts $\{Pt^{+}, Pt^{-}\}$. These logits measure the similarity between the image and each prompt. Assuming $z^{pre(+)}_{i, j}$ represents the logits for all positive prompts $Pt^{+}$ given the pre-disaster image $x^{pre}_{i, j}$, and the same applies to other combinations. Then, the logits from the CLIP model can be calculated as:
\begin{equation}
\begin{split}
z^{pre(+)}_{i, j} &= \text{softmax}\bigl(E_I(x^{pre}_{i, j}) \cdot (E_T(Pt^{+}))^\top\bigr) \\
% z^{pre(-)}_{i, j} &= \text{softmax}\bigl(E_I(x^{pre}_{i, j}) \cdot (E_T(Pt^{-}))^\top\bigr) \\
z^{post(+)}_{i, j} &= \text{softmax}\bigl(E_I(x^{post}_{i, j}) \cdot (E_T(Pt^{+}))^\top\bigr) \\
z^{post(-)}_{i, j} &= \text{softmax}\bigl(E_I(x^{post}_{i, j}) \cdot (E_T(Pt^{-}))^\top\bigr)
\end{split}
\end{equation}

Next, we find the maximum value for each set of logits to be the representative confidence score and calculate the score difference between pre- and post-disaster images in positive prompts $\Delta z^{+}_{i, j}$, as well as the score difference between positive and negative prompts in post-disaster image $\Delta z^{post}_{i, j}$:
\begin{equation}
\label{eqn:zplus}
\Delta z^{+}_{i, j} = \max{z^{post(+)}_{i, j}} - \max{z^{pre(+)}_{i, j}} + \epsilon
\end{equation}
\begin{equation}
\label{eqn:zpost}
\Delta z^{post}_{i, j} = \max{z^{post(+)}_{i, j}} - \max{z^{post(-)}_{i, j}}
\end{equation}

\noindent where $\epsilon$ is a tolerance value set at 0.01 to counter minor distribution shift of bi-temporal images. Here, both $\Delta z^{+}_{i, j}$ and $\Delta z^{post}_{i, j}$ are useful indicators of building damage. The lower the values, the more likely the building has been damaged. Finally, we compute the weighted score $s_{i, j}$ from (\ref{eqn:zplus}) and (\ref{eqn:zpost}), and get the final prediction of building damage $\tilde{y}_{i, j}$:
\begin{equation}
s_{i, j} = 0.5 \times \Delta z^{+}_{i, j} + 0.5 \times \Delta z^{post}_{i, j}
\end{equation}
\begin{equation}
\tilde{y}_{i, j} = \begin{cases}
1 \text{ (undamaged)}, & \text{if } s_{i, j} \geq \tilde{\sigma}, \\
2 \text{ (damaged)}, & \text{if } s_{i, j} < \tilde{\sigma}.
\end{cases}
\end{equation}

\noindent where $\tilde{\sigma}$ is the threshold to be considered undamaged.

Lastly, the evaluation mask $\tilde{Y}_{i, j} \in \{0, 1, 2\}^{H \times W}$ of the building is computed by multiply the segmentation mask $\hat{Y}_{i, j} \in \{0, 1\}^{H \times W}$ of the building with its damage classification $\tilde{y}_{i, j}$:
% H x W is correct here
\begin{equation}
\tilde{Y}_{i, j} = \hat{Y}_{i, j} \times \tilde{y}_{i, j}
\end{equation}

To get a final evaluation mask $\tilde{Y}_i$ for the whole image pair $\{x^{pre}_i, x^{post}_i\}$, we merge all building segmentation masks of the image $\{\hat{Y}_{i, j}\}^{J_i}_{j = 1}$ pixel-wise with the largest value.

\section{U-BDD++: An Improved Approach}
\label{sec:U-BDD++}

% Fig 3: U-BDD++
\begin{figure*}[ht]
    \centering
    \includegraphics[width=1.0\textwidth]{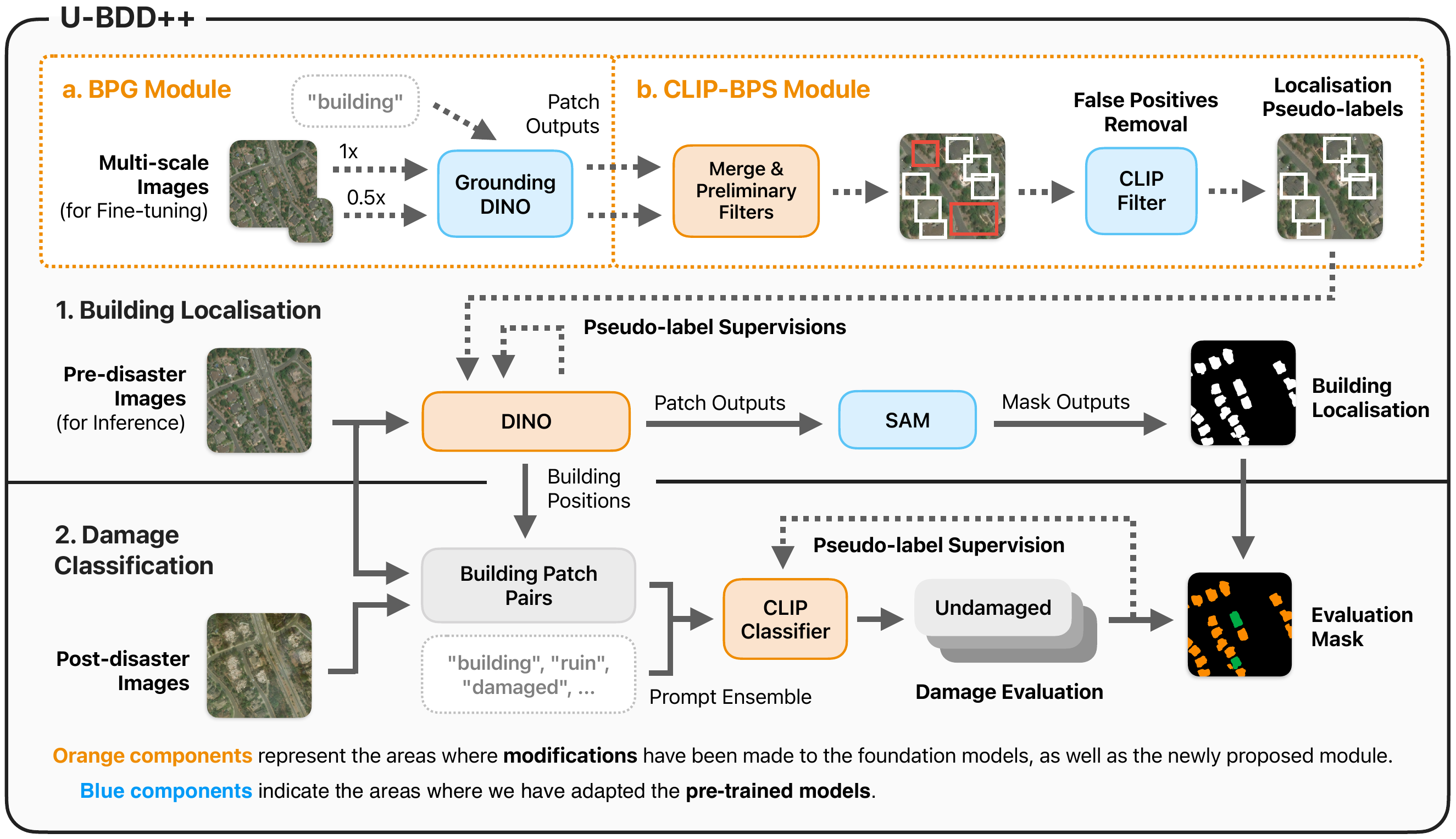}
    \captionsetup{width=0.96\linewidth}
    \caption{\textbf{The proposed workflow of U-BDD++.} U-BDD++ extends the U-BDD baseline to include fine-tuning of the foundation models in both stages. To address the domain shift issue in the satellite imagery, two specifically designed modules, BPG and CLIP-BPS, provide the model in the building localisation stage with high-quality initial fine-tuning supervision.}
    \label{fig:U-BDD++}
\end{figure*}

\subsection{Overview}

Despite the competitive performance of the U-BDD baseline in identifying and determining the damages to the buildings, it is constrained by a significant domain gap between the data used to pre-train the foundational models and the satellite images. By observing the predictions from the U-BDD baseline, we found that this domain gap exists due to: 1) The sizes of buildings in this task vary considerably compared to generic object detection tasks due to the limited resolution of satellite images. Small buildings can appear as tiny as a few pixels in the image, and foundation models pre-trained on various common object detection tasks may have low confidence values on these buildings. 2) The top-down view characteristic of satellite imagery often leads to visual similarities among objects with comparable semantic attributes. For instance, man-made structures like parking lots and tennis courts could have similar structure outlines when observed from a satellite, making Grounding DINO~\cite{gdino} struggle to effectively distinguish between them. Consequently, this diminishes the reliability and robustness of the outcomes.

To better adapt to the domain of satellite images in unsupervised settings, we propose \textbf{U-BDD++}, an improved framework built on the U-BDD baseline to enhance the performance on both building localisation and damage classification stages. The framework overview is depicted in Fig.~\ref{fig:U-BDD++}. Considering the significance of the building localisation stage, which directly influences the subsequent damage classification performance, and the distinct prominence of the two domain gap issues at this stage, we have developed two innovative modules. These are specifically designed to better guide the foundation model during the process of building localisation. The \textbf{Building Proposal Generation (BPG)} module solves the first domain gap issue of building sizes by converting the image into multiple patches with different scales and locations to guide the model to predict more building-like objects. The \textbf{CLIP-enabled noisy Building Proposal Selection (CLIP-BPS)} module solves the second domain gap issue of similar semantics by reusing the pre-trained CLIP~\cite{clip} in the second stage to also discriminate against false positive proposals from the BPG modules. Additionally, we adopt a pseudo-labelling strategy to fine-tune the foundation models in both stages. The initial pseudo-labels for building localisation are derived from the BPG and CLIP-BPS modules, while subsequent fine-tuning of building localisation and damage classification leverage pseudo-labels produced directly by the foundation models themselves.

% Fig 4: CLIP-BPS
\begin{figure*}[ht]
\centering
    \includegraphics[width=1\textwidth]{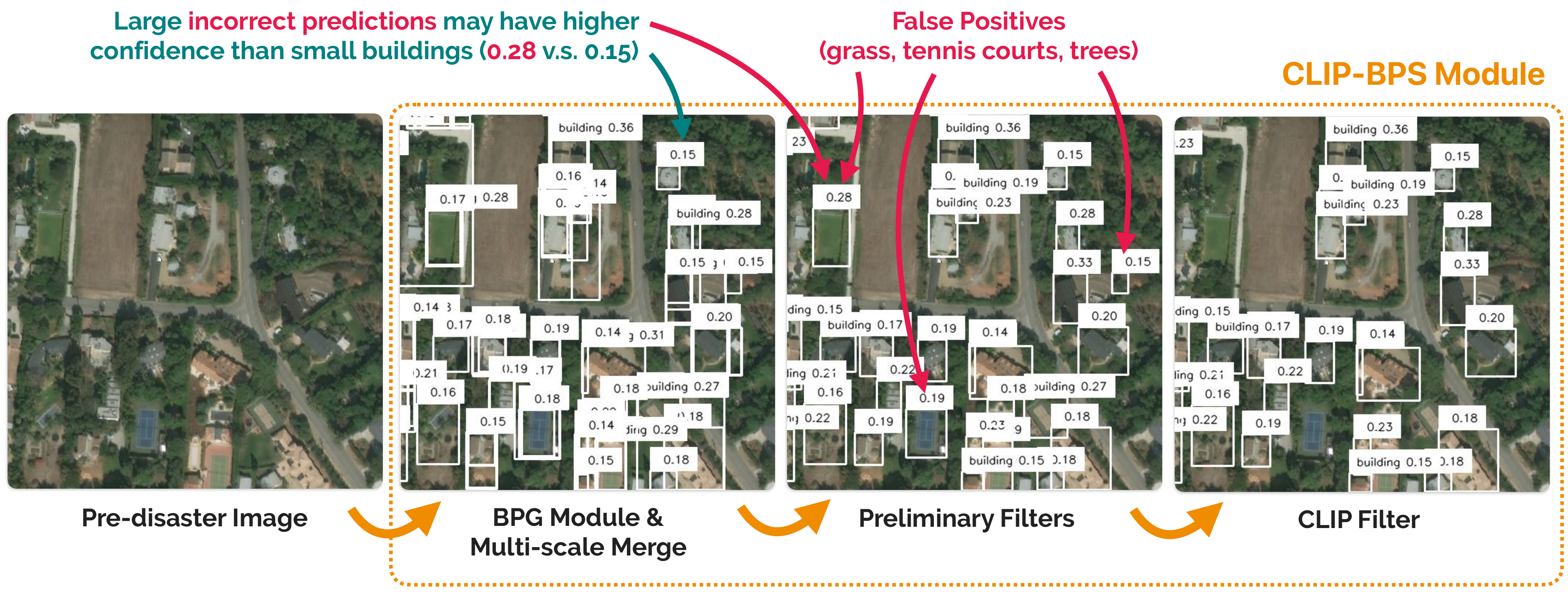}
    \captionsetup{width=0.96\linewidth}
    \caption{\textbf{Example visual demonstration of CLIP-BPS module.} Multiscale merging and preliminary filters remove duplicate or incorrectly sized bounding boxes from the BPG module, while preserving most buildings. Both merging and filtering can be processed simultaneously. For visualisation clarity, multiscale merge is processed first before preliminary filters. Finally, the CLIP filter removes false positives and large incorrect predictions with similar semantic traits, such as tennis courts.}
\label{fig:CLIP-BPS}
\end{figure*}

\subsection{BPG Module}
\label{sebsec:BPG}

The BPG module is the first part of generating pseudo-labels for the initial building localisation supervision, as shown in Fig.~\ref{fig:U-BDD++}(a). It consists of a multiscale splitting step for converting images into multiple scales, and a Grounding DINO~\cite{gdino} prediction step to generate all possible building bounding boxes in different sizes.

\medskip
\textbf{Multiscale Splitting.} Given a set of $N'$ unlabelled training images $\{x^{\prime}_{i}\}^{N^{\prime}}_{i=1}$, to handle buildings of various scales and sizes in the images, we follow the idea in~\cite{bddmtf} to divide each image into smaller patches. Each patch is initially sized at $(\lambda H \times \lambda W)$, where $\lambda \in (0, 1]$ is the scale relative to the original image dimensions. For calculation convenience, each smaller $\lambda$ is set at half of the previous scale, \eg, $1, 0.5, \ldots, \lambda_{n}$, where n is the total number of scales. Apart from~\cite{bddmtf}, we further use a sliding window technique to make patches to have overlaps to minimise the situation where buildings are located at the edge of a patch and therefore not detected. Assuming the stride for each patch is exactly half the width of the patch, the total number of smaller patches at scale $\lambda$ is:
$\left\lceil \frac{H}{0.5 \lambda H} - 1 \right\rceil \times \left\lceil \frac{W}{0.5 \lambda W} - 1 \right\rceil = \left\lceil \frac{2}{\lambda} - 1 \right\rceil \times \left\lceil \frac{2}{\lambda} - 1 \right\rceil$.
% \begin{equation}
% \left\lceil \frac{H}{0.5 \lambda H} - 1 \right\rceil \times \left\lceil \frac{W}{0.5 \lambda W} - 1 \right\rceil = \left\lceil \frac{2}{\lambda} - 1 \right\rceil \times \left\lceil \frac{2}{\lambda} - 1 \right\rceil
% \end{equation}
Both the original image and its smaller patches are processed in the next step.

\medskip
\textbf{Grounding DINO Prediction.} We adopt a multiscale image prediction strategy, allowing Grounding DINO~\cite{gdino} to generate a larger quantity of building proposals. The smaller patches are firstly bi-linearly up-sampled to the same dimension to ensure consistent logit outputs. Then, for each image patch, we utilise the pre-trained Grounding DINO~\cite{gdino} with text prompt ``building'' to generate bounding boxes for building predictions. We denote the set of $J^{\prime}_{i, \lambda}$ bounding boxes generated for all image patches at scale $\lambda$ as:
\begin{equation}
\label{eqn:bprimeilambda}
B^{\prime}_{i, \lambda} = \{(x_j, y_j, h_j, w_j) \mid j=1, 2, \ldots, J^{\prime}_{i, \lambda} \}
\end{equation}

\noindent where $x_j, y_j, h_j, w_j$ represent the left, top, height and width of the bounding boxes, and are relative to the original image dimensions $(H \times W)$. Each bounding box will also have its corresponding logit outputted from Grounding DINO, and we denote the logit set at scale $\lambda$ as:
\begin{equation}
\label{eqn:zprimeilambda}
Z^{\prime}_{i, \lambda} = \{z_j \mid j=1, 2, \ldots, J^{\prime}_{i, \lambda} \}
\end{equation}

In practice, we set a lower box threshold $\sigma^{\prime}_{G} \leq \sigma_{G}$ to mitigate the difficulties of locating smaller buildings. The resulting bounding boxes and the logits will be passed to the next module for further processing.

\subsection{CLIP-BPS Module}
\label{subsec:CLIP-BPS}

To solve the domain gap issue of similar semantics, we propose a bounding box filtering module after the BPG module, named CLIP-BPS, as the second part of the pseudo-label generation process. The overview pipeline is shown in Fig.~\ref{fig:U-BDD++}(b). CLIP-BPS removes inaccurate predictions while preserving most of the buildings at different scales. The workflow consists of two steps: 1) Multi-scale Merging and Preliminary Filtering, and 2) CLIP Filtering. To visualise the process, Fig.~\ref{fig:CLIP-BPS} provides a clear demonstration of each step in the CLIP-BPS module.

\medskip
\textbf{Multiscale Merging and Preliminary Filtering.} Given the fact that satellite images are taken from a certain height from Earth, we empirically find that the large predicted bounding boxes often contain more than one building, which hinders the prediction performance. As a remedy to this issue, we first aim to remove the bounding boxes $B^{\prime}_{i, \lambda}$ from (\ref{eqn:bprimeilambda}). Specifically, to obtain the filtered set of bounding boxes $B^{F}_{i, \lambda}$, we further reject the bounding box proposal based on the maximum size, width, and height criteria. Note the filters are relative to the scale of the patch $(\lambda H \times \lambda W)$. As such, large buildings in small patches may be filtered out, but can still remain in larger patches or in the original image. We argue that these criteria can be easily obtained from the satellite specifications. Similar to (\ref{eqn:zprimeilambda}), we can obtain the corresponding filtered logits $Z^{F}_{i, \lambda}$.

% with an area close to the entire patch size, or the width or height close to the patch width and height respectively, as they usually indicate a full-patch prediction. We define a filtered set of bounding boxes after dimension filters, $B^{F}_{i, \lambda}$, and the corresponding filtered logits $Z^{F}_{i, \lambda}$ as follows:
% \begin{equation}
% \begin{split}
% % B^{F}_{i, \lambda} & = \{(x_j, y_j, h_j, w_j) \in B^{\prime}_{i, \lambda} \mid h_j \leq \sigma_{H}(\lambda H) \text{ and} \\
% % & w_j \leq \sigma_{W}(\lambda W) \text{ and } h_j \times w_j \leq \sigma_{A}(\lambda H \lambda W) \}
% B^{F}_{i, \lambda} & = \{(x_j, y_j, h_j, w_j) \mid (x_j, y_j, h_j, w_j) \in B^{\prime}_{i, \lambda} \text{ and } \\
% & h_j \leq \sigma_{H}(\lambda H) \text{ and } w_j \leq \sigma_{W}(\lambda W) \text{ and} \\
% & h_j \times w_j \leq \sigma_{A}(\lambda H \lambda W) \}
% \end{split}
% \end{equation}
% \begin{equation}
% \begin{split}
% Z^{F}_{i, \lambda} & = \{b_j \mid b_j \in Z^{\prime}_{i, \lambda} \text{ and } j \in \{j \mid \\
% & (x_j, y_j, h_j, w_j) \in B^{F}_{i, \lambda}\}\}
% \end{split}
% \end{equation}

% \noindent where $\sigma_{H}, \sigma_{W}, \sigma_{A}$ are height, width and area threshold coefficients, respectively. Note the filters are relative to the scale of the patch $(\lambda H \times \lambda W)$. As such, large buildings in small patches may be filtered out, but can still remain in larger patches or in the original image.

As the same building may have multiple overlapping bounding boxes due to predictions from different scales, we further apply a Non-Maximum Suppression (NMS) filter to $B^{F}_{i, \lambda}$ with a fixed IoU threshold of 0.1. This keeps the bounding boxes of the highest logits from Grounding DINO~\cite{gdino}. We denote the set of remaining boxes after the NMS filter as $B^{N}_{i, \lambda}$. Meanwhile, all bounding boxes of all scales $\lambda = \{1, 0.5, \ldots, \lambda_{n}\}$ are merged into a single set $B^{\prime\prime}_i$:
\begin{equation}
\label{eqn:bprimeprime}
B^{\prime\prime}_i = B^{N}_{i, 1} \cup B^{N}_{i, 0.5} \cup \ldots \cup B^{N}_{i, \lambda_{n}}
\end{equation}

\medskip
\textbf{CLIP Filtering.} Finally, a CLIP~\cite{clip} filter is applied to validate whether the object inside the bounding box is indeed building. The purpose of the CLIP filter is to further remove inaccurate bounding box predictions, like false positives with similar semantics. To achieve this, we designed two criteria for determining a building: 1) Based on the confidence given a building-related prompt, and 2) Based on the comparison of probabilities among the building-related prompt and other prompts with similar semantics.

We first extract the image patch from the original image for each bounding box in (\ref{eqn:bprimeprime}) as the image input, represented as $x^{\prime}_{i, j}$. Let $Pt^{\prime}$ denote a list of $K$ prompts, with the first prompt $Pt^{\prime}_0$ being the building-related prompt, such as ``A satellite photo of a building''. The remaining prompts in the list $[Pt^{\prime}_1, Pt^{\prime}_2, \ldots, Pt^{\prime}_K]$ indicate similar-semantic phrases, like ``A satellite photo of a swimming pool''. The choice of prompts will be discussed in Section \ref{subsec:ExperimentBL}. When the given bounding box contains a building, CLIP~\cite{clip} should present a high enough confidence with $Pt^{\prime}_0$, and the confidence of $Pt^{\prime}_0$ should also be the highest among other prompts, corresponding to the two criteria respectively.

We reuse the image encoder $E_I$ and text encoder $E_T$ of CLIP~\cite{clip} from the U-BDD baseline. For each image patch, we use $Pt^{\prime}$ as a text prompt to CLIP, and obtain the logits $Z^{C}_{i, j} = \{z^{C}_{i, j, k}\}^{K}_{k=0}$ by multiplying the encoded image with the text vectors of the prompts:
\begin{equation}
\begin{split}
Z^{C}_{i, j} &= E_I(x^{\prime}_{i, j}) \cdot (E_T(Pt^{\prime}))^\top
\end{split}
\end{equation}

Let $z^{C}$ be the threshold for filtering the bounding boxes based on the CLIP~\cite{clip} model logits. If $z^{C}_{i, j, 0} \geq z^{C}$, we keep the bounding box; otherwise, it gets filtered. Additionally, if the maximum logit position is not related to the first building prompt, the bounding box is also removed. We can define the final set of bounding boxes $B^{\prime\prime\prime}_i$ after the filtering process as follows:
\begin{equation}
\begin{split}
B^{\prime\prime\prime}_i &= \{(x_j, y_j, h_j, w_j) \mid (x_j, y_j, h_j, w_j) \in B^{\prime\prime} \text{ and } \\
& j \in \{ j \mid z^{C}_{i, j, 0}\geq z^{C} \text{ and } \argmax_{m} z^{C}_{i, j, m} = 0\} \}
\end{split}
\end{equation}

Finally, the pipeline exports the resulting bounding boxes $B^{\prime\prime\prime}_i$ for each image $x^{\prime}_{i}$ as pseudo-labels for fine-tuning DINO.

\subsection{Self-training Module}
\label{subsec:ModelSelf-training}
\medskip
In the building localisation stage, we directly fine-tune the DETRs~\cite{detr}, such as DINO~\cite{dino}. The initial supervision is generated from the BPG module and CLIP-BPS module. The subsequent fine-tuning is done by using the DINO model itself, which chooses the bounding boxes with the highest confidence as pseudo-labels. In the damage classification stage, after the initial prediction of the building damage for all training data using the U-BDD baseline, CLIP~\cite{clip} chooses the predictions with the highest confidence for both damaged and undamaged types and uses them for the supervision to further fine-tune the model itself, similar to the DINO~\cite{dino} fine-tuning in the building localisation stage. To avoid the catastrophic forgetting issue presented in fine-tuning CLIP, we freeze both the text encoder and image encoder of CLIP, and add an adaptor to the encoding states for training the model. Finally, the cross-entropy loss is used given the pseudo-labels generated.

\section{Experiment}
\label{sec:Experiment}

\subsection{Dataset}
\label{subsec:Setup}
We use the public xBD dataset~\cite{xbd-dataset, xbd-dataset-2} for our experiment. xBD contains high-resolution $1024\times1024$ images with over 850,000 building annotations under a range of disasters including floods, wildfires, \etc. In our experiment, we use only the standard training set and test set provided~\cite{xbd-dataset-link}. Although the dataset provides more than one level of damage, Shen \etal~\cite{bdanet-2, bdanet} suggest there are difficulties in distinguishing certain types of damage under fully-supervised settings, such as between minor and major damage.
% One possible explanation is the similarities between the two damage types when viewed from a top-down perspective.
For our unsupervised method, we simplify the evaluation to only consider \emph{damaged} and \emph{undamaged} categories. Specifically, \emph{no damage} in xBD is considered as \emph{undamaged}, while \emph{minor damage}, \emph{major damage} and \emph{destroyed} are all considered as \emph{damaged}.

% One possible explanation is that the definition of minor damage also cooperates with non-structural straits such as water and volcanic flow nearby~\cite{xbd-dataset-2}. Minor damage and major damage, on the other hand, can also have negligible visual differences~\cite{xbd-dataset}.
% on out-of-distribution predictions? different types of disasters

\subsection{Building Localisation}
\label{subsec:ExperimentBL}
\textbf{Implementation.} In the U-BDD baseline, we use Grounding DINO~\cite{gdino} pre-trained on generic object detection datasets such as COCO~\cite{COCO} with Swin-T~\cite{swin} backbone to perform building detection; In U-BDD++, we fine-tune a pre-trained DINO~\cite{dino} model with both ResNet-50~\cite{resnet} and Swin-L~\cite{swin} backbones using 2 NVIDIA GeForce RTX 3090 Ti GPUs. Both learning rate and weight decay are set to 1e-4, and the models are trained for 12 epochs with a batch size of 1. We denote the localisation model with box threshold $\sigma_{G}$ for predictions as ``[model]@[threshold]'', \eg~Grounding DINO@0.35. We compare Grounding DINO@0.35, Grounding DINO@0.15 and DINO Fine-tuned@0.35 (2 backbones), and report their corresponding AP values object-wise. For the segmentation, we use pre-trained SAM~\cite{sam} with ViT-H~\cite{vit} backbone and report the F1 and IoU values pixel-wise, where buildings are positive and backgrounds are negative. The overall results are shown in Table~\ref{tab:building-seg}.

\textbf{U-BDD++ Modules.} For BPG module in U-BDD++, we use 2-scale image splitting with $\lambda$ set to 1 and 0.5 for main comparison. The building proposals are generated by pre-trained Grounding DINO@0.14~\cite{gdino}. We additionally conduct 3 settings with different $\lambda$s: 1) 1x, 2) 1x and 0.5x, 3) 1x, 0.5x and 0.25x, and fine-tune ResNet-50~\cite{resnet} DINO~\cite{dino} models with the pseudo-label generated (the other U-BDD++ components are fixed). The results are reported in F1 and IoU with SAM~\cite{sam} applied, and are shown in Table \ref{tab:ablation-scale}. In the CLIP-BPS module, the preliminary filters $\sigma_{H}, \sigma_{W} \text{ and } \sigma_{A}$ are set to 0.75, 0.75 and 0.03, respectively. The CLIP filter uses pre-trained CLIP~\cite{clip} with ViT-L/14@334px~\cite{vit} backbone for filtering out false-positives. The prompt ensemble uses 10 prompts in the format of ``A satellite photo of [object]'', where [object] is one of \{building, swimming pool, tennis court, parking lot, street, trees, grass, soil, car, truck\}. To ensure CLIP has enough context when making the prediction, we apply padding to each patch with surrounding pixels of 10 in each dimension, or until the padded patch reaches at least 50x50.

% Result: Building localisation/segmentation
\begin{table*}[ht]
\centering
\captionsetup{width=1\linewidth}
\caption{\textbf{Results on the building localisation and segmentation.} F1\textsubscript{SAM} and IoU\textsubscript{SAM} indicate the use of pre-trained SAM immediately after the bounding box outputs from the localisation models and are measured pixel-wise. AP scores evaluate the localisation models only and are measured object-wise. The first row represents the upper bound where the localisation model is fine-tuned on the ground truth label in a fully-supervised manner, while the rest are self-supervised using the label source from either baseline or U-BDD++.}
\begin{tabular}{l l l c c c c c c c c}
\toprule
    Label Source & Localisation Model & Backbone & F1\textsubscript{SAM} & IoU\textsubscript{SAM} & AP & AP\textsubscript{50} & AP\textsubscript{75} & AP\textsubscript{small} & AP\textsubscript{med} & AP\textsubscript{large}  \\
\midrule
    Ground Truth & DINO Fine-tuned@0.35~\cite{dino} & Swin-L~\cite{swin} & 81.2 & 71.7 & 25.5 & 41.2 & 27.7 & 14.3 & 39.2 & 43.5 \\
\midrule
    U-BDD Baseline & Grounding DINO@0.35~\cite{gdino} & Swin-T & 39.9 & 33.4 & 1.8 & 3.0 & 1.6 & 0.6 & 3.2 & 8.9  \\
    U-BDD Baseline & Grounding DINO@0.15 & Swin-T & 43.8 & 34.3 & 4.7 & 10.3 & 3.7 & 1.7 & 9.5 & 6.5 \\
    U-BDD++ (ours) & DINO Fine-tuned@0.35 & RN-50~\cite{clip} & 55.9 & 46.3 & 5.4 & 11.9 & 4.2 & 1.3 & 10.4 & 10.7 \\
    U-BDD++ (ours) & DINO Fine-tuned@0.35 & Swin-L & \textbf{58.2\tiny{$\pm$1.2}} & \textbf{47.4\tiny{$\pm$1.1}} & \textbf{6.4\tiny{$\pm$0.6}} & \textbf{13.7\tiny{$\pm$1.2}} & \textbf{5.2\tiny{$\pm$0.6}} & \textbf{1.9\tiny{$\pm$0.2}} & \textbf{12.1\tiny{$\pm$1.2}} & \textbf{11.8\tiny{$\pm$0.1}} \\
    % backup %
    % U-BDD++ (ours) & DINO Fine-tuned@0.35 & RN-50~\cite{clip} & 55.2 & 44.4 & 6.4 & 12.9 & 5.7 & 1.3 & 12.2 & \textbf{15.4} \\
    % U-BDD++ (ours) & DINO Fine-tuned@0.35 & Swin-L & \textbf{59.3} & \textbf{48.4} & \textbf{7.0} & \textbf{14.9} & \textbf{5.8} & \textbf{2.1} & \textbf{13.3} & 11.8 \\
\bottomrule
\end{tabular}
\label{tab:building-seg}
\end{table*}

% Result: Damage classification
\begin{table*}[ht]
\centering
\captionsetup{width=1\linewidth}
\caption{\textbf{Results on the building damage classification.} The \emph{Building Localisation Data} column specifies whether the data source of building bounding boxes comes from ground truth, the first stage of baseline, or from the first stage of U-BDD++ (end-to-end building localisation and damage classification of U-BDD++). The \emph{CLS Label Source} column specifies whether damage classification is performed by using the baseline method, or the self-supervised adaptor method. ($^\dag$Low building predictions can result in high F1\textsubscript{B} values.)}
% *Calculated by the mean of all damage-type results.}
\begin{tabular}{l l l l c c c c c c c c}
\toprule
Building Localisation Data & CLS Label Source & CLS Model & Backbone & F1\textsubscript{B} & F1\textsubscript{U} & F1\textsubscript{D} & mF1 & IoU\textsubscript{B} & IoU\textsubscript{U} & IoU\textsubscript{D} & mIoU \\
\midrule
    % BDANet~\cite{bdanet} & Ground Truth & BDANet~\cite{bdanet} & RN-50~\cite{resnet} & 86.4 & 92.5 & 76.0* & 85.0 & - & - & - & -\\
    Ground Truth & U-BDD Baseline & CLIP~\cite{clip} & ViT-L~\cite{vit} & 99.2 & 64.2 & 49.1 & 70.8 & 98.4 & 57.9 & 44.5 & 66.9 \\
    % backup
    % Ground Truth & U-BDD Baseline & CLIP~\cite{clip} & ViT-L/14@336px & 74.7 & 83.4 & - & - \\
\midrule
    U-BDD Baseline First-Stage & U-BDD Baseline & CLIP & ViT-L & 93.4$^\dag$ & 39.5 & 34.1 & 55.7 & 88.8$^\dag$ & 32.6 & 31.0 & 50.8 \\
    U-BDD++ First-Stage (ours) & U-BDD Baseline & CLIP & ViT-L & \textbf{97.3}$^\dag$ & 48.9 & \textbf{42.4} & 62.9 & \textbf{95.1}$^\dag$ & 41.4 & \textbf{39.0} & 58.5 \\
    U-BDD++ First-Stage (ours) & U-BDD++ (ours) & Adapted CLIP & ViT-L & \textbf{97.3}$^\dag$ & \textbf{55.1} & 39.1 & \textbf{63.8} & \textbf{95.1}$^\dag$ & \textbf{48.0} & 34.1 & \textbf{59.1} \\
    % U-BDD Baseline First-Stage & U-BDD Baseline & CLIP & ViT-L/14 & 93.4$^\dag$ & 39.5 & 34.1 & 55.7 & 88.8$^\dag$ & 32.6 & 31.0 & 50.8 \\
    % U-BDD++ First-Stage (ours) & U-BDD Baseline & CLIP & ViT-L/14 & \textbf{97.3}$^\dag$ & 48.9 & \textbf{42.4} & 62.9 & \textbf{95.1}$^\dag$ & 41.4 & \textbf{39.0} & 58.5 \\
    % U-BDD++ First-Stage (ours) & U-BDD++ (ours) & CLIP+Adaptor & ViT-L/14 & \textbf{97.3}$^\dag$ & \textbf{55.1} & 39.1 & \textbf{63.8} & \textbf{95.1}$^\dag$ & \textbf{48.0} & 34.1 & \textbf{59.1} \\

    % backup
    % U-BDD Baseline First-Stage & U-BDD Baseline & CLIP & ViT-L/14@336px & - & - & 26.2 & 22.2 \\
    % U-BDD++ First-Stage (ours) & U-BDD Baseline & CLIP & ViT-L/14@336px & - & - & 44.1 & 41.0 \\
    % U-BDD++ First-Stage (ours) & U-BDD++ (ours) & CLIP + Adaptor & ViT-L/14@336px & - & - & - & - \\
\bottomrule
\end{tabular}
\label{tab:building-dmg}
\end{table*}

% Result: BPG Module Scale
\begin{table}[ht]
\centering
\caption{\textbf{Effects on the multiscale splitting in BPG module.} Pre-trained SAM with ViT-H is applied after bounding box prediction.}
\begin{tabular}{c c c c c}
\toprule
Scales $\lambda$ & Patches & Model & F1\textsubscript{SAM} & IoU\textsubscript{SAM} \\
\midrule
1.0 & 1 & DINO@0.35~\cite{dino} & 52.5 & 42.7 \\
1.0 \& 0.5 & 10 & DINO@0.35 & \textbf{55.2} & \textbf{44.4} \\
1.0 \& 0.5 \& 0.25 & 59 & DINO@0.35 & 53.7 & 43.6 \\
\bottomrule
\end{tabular}
\label{tab:ablation-scale}
\end{table}

\textbf{Results.} As shown in Table~\ref{tab:building-seg}, DINO Fine-tuned@0.35~\cite{dino} with ResNet-50~\cite{resnet} backbone out-performs baseline Grounding DINO@0.35~\cite{gdino} with Swin-T~\cite{swin} backbone by 3.6\% in AP and 8.9\% in AP\textsubscript{50}, showing the performance increase of DINO with U-BDD++ fine-tune method. DINO Fine-tuned@0.35~\cite{dino} with Swin-L~\cite{swin} backbone additionally boosts the performance by a large margin of 18.3\% in F1\textsubscript{SAM} and 14\% in IoU\textsubscript{SAM} compared to Grounding DINO@0.35. This shows the effectiveness and robustness of U-BDD++ in building localisation. We additionally provide a DINO~\cite{dino} fine-tuned with ground truth labels in the first row of Table~\ref{tab:building-seg} as our upper-bound results. With the introduction of U-BDD++, the self-training result is closing the gap between itself and the upper bound, demonstrating the competitiveness of U-BDD++ and the potential of U-BDD.

The visual comparisons between baseline and U-BDD++ are shown in Fig.~\ref{fig:VisualResults}, where both of our methods produce robust and adequate building predictions. In the first two rows, the U-BDD++ framework can detect more buildings under different scales than the baseline, demonstrating the effectiveness of the BPG module. U-BDD++ can also detect fewer false positives than baseline, as shown in the top-right corners in the second-row images and the bottom-left corners in the third-row images, showing the effectiveness of CLIP-BPS modules.

We also justify the choice of scales used for multiscale prediction of the BPG module. As shown in Table~\ref{tab:ablation-scale}, the 2-scale setting (1x and 0.5x) outperforms the single scale with the original image (1x) and the three-scale setting (1x, 0.5x and 0.25x). The three-scale setting may perform worse than the two-scale setting due to the additional noise introduced at very small scales, which the preliminary filter might not be able to detect. Additionally, predicting on the tiny pixels could pose a challenge for the CLIP~\cite{clip} model.

\subsection{Damage Classification}
\textbf{Implementation.} In the U-BDD baseline, we use prompt ensemble with $Pt^{+} =$ \{``A satellite photo of a building'', ``normal building'', ``undamaged building'', ``building''\}, and $Pt^{-} =$ \{``A satellite photo of a ruin'', ``damaged building'', ``destroyed building'', ``ruin''\} for CLIP prediction. Each patch is padded with the same configuration of building localisation before feeding to the CLIP model. In U-BDD++, we adapt a pre-trained CLIP~\cite{clip} model with ViT-L/14@334px~\cite{vit} backbone by adding MLP layers to the encoding space while freezing both image and text encoders of CLIP for training. To train the model, we obtain the top 10\% of the baseline predictions in confidence for both damaged and undamaged as labels and use cross-entropy for calculating the loss. The training uses Adam~\cite{adam} as the optimiser, with a learning rate of 1e-4, weight decay of 5e-5, and a batch size of 16 for 12 epochs. For both methods, we report F1 and IoU scores for the backgrounds (F1\textsubscript{B}/IoU\textsubscript{B}), undamaged buildings (F1\textsubscript{U}/IoU\textsubscript{U}) and damaged buildings (F1\textsubscript{D}/IoU\textsubscript{D}) pixel-wise. We additionally evaluate means of F1 and IoU scores (mF1/mIoU) across all types of damages and the background. The complete results are shown in Table \ref{tab:building-dmg}. In the results, we include all possible building patch sources for CLIP, ranging from ground truth, GroundingDINO@0.35 from baseline and DINO Fine-tuned@0.35 from U-BDD++ (\ie, end-to-end damage detection).
% We also include BDANet~\cite{bdanet} as the current SoTA in supervised building damage detection for reference.

\textbf{Results.} As shown in Table~\ref{tab:building-dmg}, both baseline and U-BDD++ achieve competitive overall F1\textsubscript{U} and F1\textsubscript{D} scores, closing the gap to the upper-bound method. This shows the high performance in damage classification tasks in general, and the effectiveness of prompt ensemble techniques. Additionally, the adapted CLIP in U-BDD++ outperforms the baseline CLIP method by 0.8\% in mF1, and 0.6\% in mIoU, demonstrating the benefits of self-training in U-BDD++.

\begin{figure*}[ht!]
    \centering
    \includegraphics[width=1.0\textwidth]{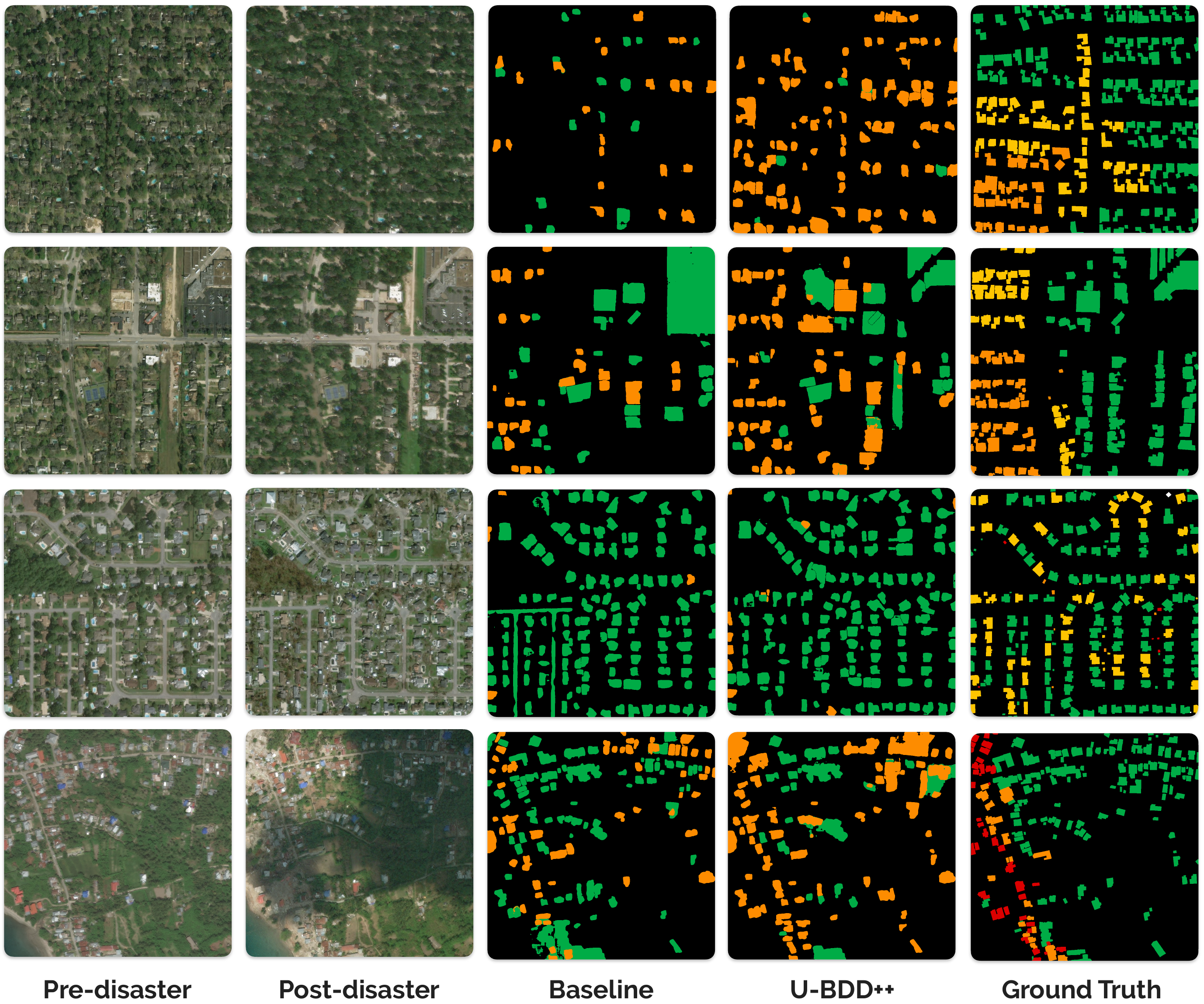}
    \captionsetup{width=0.96\linewidth}
    \caption{\textbf{Visual predictions from U-BDD baseline and U-BDD++ on end-to-end building localisation and damage classification.} Each column represents the pre-disaster images, the post-disaster images, the baseline evaluation mask, the U-BDD++ evaluation mask and the ground truth evaluation mask respectively. Baseline and U-BDD++ classify buildings as undamaged (green) and damaged (orange). Ground truth masks further classify buildings as no damage (green), minor damage (yellow), major damage (orange) and destroyed (red). Note that the different damage levels in the ground truth masks are for reference only, and all damage levels except no damage will be considered damaged in U-BDD. Both pre-trained Grounding DINO in baselines and the fine-tuned DINO in U-BDD++ have a prediction box threshold $\sigma_{G}$ of 0.15.}
    \label{fig:VisualResults}
\end{figure*}

In the visual comparisons (Fig.~\ref{fig:VisualResults}), a majority of the building damages are correctly classified by both the baseline and U-BDD++. However, U-BDD++ demonstrates a superior ability to accurately classify building damages, as evident, particularly in row 4 where most buildings on the left side are correctly classified as damaged. This clearly underscores the potential of utilising the pre-trained model predictions in U-BDD++.

% Fig 5: Baseline vs U-BDD++
% \begin{figure*}[ht!]
%     \centering
%     \includegraphics[width=1.0\textwidth]{figures/Visual Comparison (medium).pdf}
%     \captionsetup{width=0.96\linewidth}
%     \caption{\textbf{Visual predictions from U-BDD baseline and U-BDD++ on end-to-end building localisation and damage classification.} Each column represents the pre-disaster images, the post-disaster images, baseline building localisation (Baseline LOC), U-BDD++ localisation (U-BDD++ LOC), baseline evaluation mask (Baseline CLS), U-BDD++ evaluation mask (U-BDD++ CLS), and ground truth evaluation mask respectively. Baseline and U-BDD++ classify buildings as undamaged (green) and damaged (orange). Ground truth masks further classify buildings as no damage (green), minor damage (yellow), major damage (orange) and destroyed (red). Note that the different damage levels in the ground truth masks are for reference only, and all damage levels except no damage will be considered damaged in U-BDD. Both pre-trained Grounding DINO in baselines and the fine-tuned DINO in U-BDD++ have a prediction box threshold $\sigma_{G}$ of 0.15.}
%     \label{fig:VisualResults}
% \end{figure*}

\subsection{Ablation Study}

\textbf{Effects on Prompt Ensemble in Damage Classification.} We demonstrate the effectiveness of the prompt ensemble strategy used as the CLIP~\cite{clip} predictor in both baseline and U-BDD++ pseudo-label generation. We use building patches directly from ground truth localisation data, allowing Precision, Recall, and F1 to measure the performance object-wise (A positive prediction indicates the building is undamaged). Three configurations are compared, including 1) the logit difference of positive- and negative-prompts in the post-disaster image, 2) the confidence difference of positive-prompt between pre- and post-disaster images, and 3) prompt ensemble. The model uses pre-trained CLIP~\cite{clip} with ViT-L/14@336px~\cite{vit} backbone, and the results are reported in Table \ref{tab:ablation-clsstrat}. In the results, although all three methods achieve high performance, the prompt ensemble has the highest F1, outperforming the post-disaster-only strategy by 5.7\% and confidence change by 4.9\%. This proves that considering both the confidence change and the categorical difference is more suitable for unsupervised damage classification.

% Ablation: Baseline CLIP Damage Classification
\begin{table}[ht]
\centering
\caption{\textbf{Comparison of damage classification strategies.} The scores are calculated object-wise given ground truth building patches.}
\begin{tabular}{c c c c c}
\toprule
Method & Precision\textsubscript{obj} & Recall\textsubscript{obj} & F1\textsubscript{obj} \\
\midrule
Post-disaster Only & 87.4 & 69.7 & 77.6 \\
Confidence Change ($\Delta z^{+}$) & 80.1 & 76.7 & 78.4 \\
Prompt Ensemble ($s$) & 84.1 & 82.5 & \textbf{83.3} \\
\bottomrule
\end{tabular}
\label{tab:ablation-clsstrat}
\end{table}

\begin{figure}[ht]
    \centering
    \includegraphics[width=0.48\textwidth]{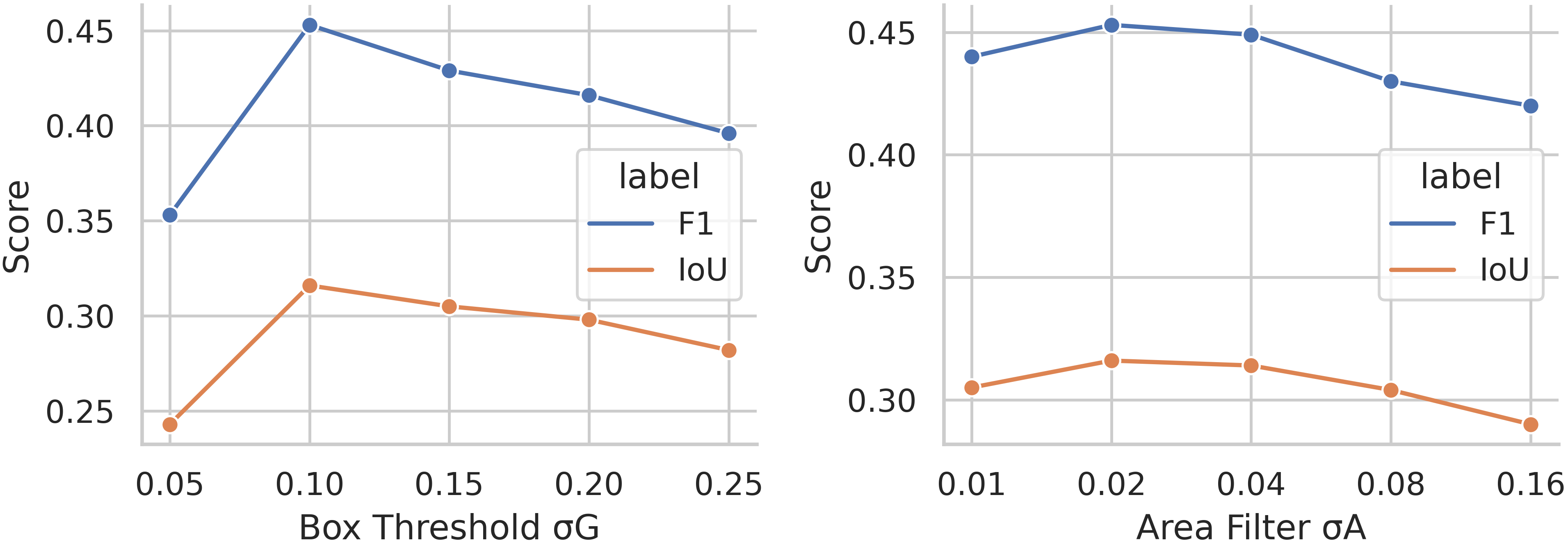}
    \captionsetup{width=0.96\linewidth}
    \caption{Comparisons of different box thresholds $\sigma_{G}$ and area filters $\sigma_{A}$ in U-BDD++.}
    \label{fig:BoxGraph}
\end{figure}

\textbf{Effects on Box Thresholds and Area Filters in U-BDD++.} We demonstrate the parameter sensitivities of the box thresholds of Grounding DINO~\cite{gdino} and scale filters (area filters in particular) in BPG and CLIP-BPS modules, respectively. We conduct the experiment by directly applying the threshold and filters to 1-scale image and apply SAM~\cite{sam} after the bounding box predictions with a ViT-H~\cite{vit} backbone for F1 and IoU evaluation. As shown in Fig. \ref{fig:BoxGraph}, box threshold $\sigma_{G} = 0.1$ and area filter $\sigma_{A} = 0.02$ achieve the highest in the F1 and IoU values, justifying the parameter choices used for DINO~\cite{dino} fine-tuning in U-BDD++.

% Fig 6: Ablation: DINO box threshold

\textbf{Effects on Filters in CLIP-BPS Module.} To validate each filter in the CLIP-BPS module, we fix the 1-scale building proposal generation from Grounding DINO@0.1~\cite{gdino} in the BPG module and apply the preliminary filters (NMS \& Scale filters) and CLIP filter in order. The bounding boxes output is applied to SAM~\cite{sam} with a ViT-H~\cite{vit} backbone for F1 and IoU evaluation. As illustrated in Table \ref{tab:ablation-clipbps}, the application of scale filters significantly enhances the F1 and IoU by 27.2\% and 20.2\%, respectively. The addition of CLIP filters further boosts the F1 and IoU by 2.6\% and 1.3\%, respectively. These results show the importance of each filter within CLIP-BPS.

% Ablation: CLIP-BPS Module
\begin{table}[ht]
\centering
\caption{Effects on each component in CLIP-BPS module.}
\begin{tabular}{l c c c c c}
\toprule
Components & Model & F1\textsubscript{SAM} & IoU\textsubscript{SAM} \\
\midrule
BPG + NMS Filter & Grounding DINO@0.1~\cite{gdino} & 15.5 & 10.1 \\
+ Scale Filter & Grounding DINO@0.1 & 42.7 & 30.3 \\
+ CLIP Filter & Grounding DINO@0.1 & \textbf{45.3} & \textbf{31.6} \\
\bottomrule
\end{tabular}
\label{tab:ablation-clipbps}
\end{table}

\section{Conclusion}
\label{sec:Conclusion}

In this paper, we propose a novel yet practical unsupervised building damage detection (U-BDD) setting, where only unlabelled pre- and post-disaster satellite images are provided for building damage detection. We construct a competitive baseline method to address U-BDD by leveraging the zero-shot inference capabilities of pre-trained foundation models. 
Built upon the baseline method, we design two modules, namely the Building Proposal Generation module and the CLIP-enabled noisy Building Proposal Selection module, to improve the localisation quality. The experimental results on the benchmark dataset confirm the effectiveness of the proposed module, which surpasses the baseline results by 18.3\% and 8.1\% on F1 in building localisation and damage classification. Our proposed method sheds light on an unsupervised approach to building damage assessment. In further work, we will investigate how to assess the severity of the damaged building without the annotated data.

\section*{Acknowledgement}
This work is supported by AAGI, Australian Research Council CE200100025 and DP230101196.

\ifCLASSOPTIONcaptionsoff
  \newpage
\fi
\bibliographystyle{IEEEtran}
\bibliography{bibtex/bib/IEEEabrv,references}

\end{document}